\documentclass{article}

\PassOptionsToPackage{numbers, compress}{natbib}
\usepackage[preprint]{neurips_2023}

\usepackage[utf8]{inputenc} 
\usepackage[T1]{fontenc}    
\usepackage{hyperref}       
\usepackage{url}            
\usepackage{booktabs}       
\usepackage{amsfonts}       
\usepackage{nicefrac}       
\usepackage{microtype}      
\usepackage{xcolor}         
\usepackage{graphicx}
\usepackage{amsmath}
\usepackage{amsfonts}
\usepackage{amssymb}
\usepackage{listings}
\usepackage{paralist}
\usepackage{multirow}
\usepackage{subcaption}
\usepackage{booktabs}
\usepackage{framed}
\usepackage{siunitx}

\definecolor{lstbackground}{rgb}{0.95,0.95,0.95}
\definecolor{lstcomment}{rgb}{0.5,0.5,0.5}
\definecolor{lstkeyword}{rgb}{0,0.5,0.5}
\definecolor{lststring}{rgb}{0.5,0,0}

\lstdefinestyle{academic}{
    language=tex,
    basicstyle=\ttfamily\footnotesize,
    backgroundcolor=\color{lstbackground},
    commentstyle=\color{lstcomment},
    stringstyle=\color{lststring},
    showstringspaces=false,
    breaklines=true,
    frame=single,
    framesep=5pt,
    xleftmargin=10pt,
    xrightmargin=10pt,
    tabsize=4,
    captionpos=b,
    breakindent=0pt,
    rulecolor=\color{black},
    escapeinside={(*@}{@*)},
}

\title{Think Outside the Code: Brainstorming Boosts\\Large Language Models in Code Generation}

\author{%
	Xin-Ye Li,
	Jiang-Tian Xue,
	Zheng Xie,
	Ming Li\\
	National Key Laboratory for Novel Software Technology\\
	Nanjing University, Nanjing, 210023, China \\
	\texttt{\{lixy,xuejt,xiez,lim\}@lamda.nju.edu.cn}
}

\begin{document}

\maketitle

\begin{abstract}
Code generation aims to automatically generate source code from high-level task specifications, which can significantly increase productivity of software engineering. Recently, approaches based on large language models (LLMs) have shown remarkable code generation abilities on simple tasks. However,  generate code for more complex tasks, such as competition-level problems, remains challenging. In this paper, we introduce \textsc{Brainstorm} framework for code generation. It leverages a \emph{brainstorming} step that generates and selects diverse \emph{thoughts} on the problem to facilitate algorithmic reasoning, where the thoughts are possible blueprint of solving the problem. We demonstrate that \textsc{Brainstorm} significantly enhances the ability of LLMs to solve competition-level programming problems, resulting in a more than 50\% increase in the pass@$k$ metrics for ChatGPT on the CodeContests benchmark, achieving state-of-the-art performance. Furthermore, our experiments conducted on LeetCode contests show that our framework boosts the ability of ChatGPT to a level comparable to that of human programmers.
\end{abstract}

\section{Introduction}

Code generation is an import field that focuses on generating high-quality, executable code from problem specification of the given task, such as natural language descriptions, input-output examples or execution traces. Developing a robust code generation system would not only enhance the productivity of experienced programmers but also alleviate the challenges of programming, making it accessible to a wider audience.

Pre-trained large language models \cite{Brown2020,Austin2021,Chen2021,Nijkamp2023,OpenAI2022,OpenAI2023} have demonstrated remarkable abilities to generate Python code that solves basic programming problems. Chen et al.~\cite{Chen2021} creates the HumanEval benchmark and evaluates the Codex model, which solves 27\% of the problems. GPT-4~\cite{OpenAI2023} achieves a pass rate of 67.0\% in a zero-shot setting with one solution sampled for each problem on the HumanEval benchmark. CodeGen~\cite{Nijkamp2023} constructs the Multi-Turn Programming Benchmark that factorize problems into multi-turn prompts and shows that providing prompts to LLMs in multi-turn fashion significantly improves the performance.

However, these approaches focus on solving problems that consist of simple problem descriptions and short solutions and previous researches~\cite{Chen2021,Li2022,Zelikman2022} has demonstrated that the performance drops dramatically as the number of generated lines increases. Solving real-world programming problems entails more than simply converting primitive instructions in natural language to code and invoking correct API calls. In contrast, solving competition-level programming problems requires understanding complicated problem specifications, leveraging algorithmic reasoning to determine solution strategies, and subsequently implementing solutions in a general-purpose programming language. Therefore, competition-level code generation techniques hold greater potential for tackling real-world programming problems.

Predominant approaches in competition-level code generation primarily involve fine-tuning the pre-trained code generation models using reinforcement learning, such as AlphaCode~\cite{Li2022}, CodeRL~\cite{Le2022}. While AlphaCode achieves beginner-level performance on simulated programming competitions on the popular Codeforces\footnote{https://codeforces.com/} platform, these approaches still rely on large-scale sampling, filtering and clustering to obtain a small portion of solutions that compile and execute correctly on public test cases. However, conducting large-scale sampling in real-world scenarios is infeasible due to its high cost, making it an unaffordable approach. Additionally, a comprehensive collection of test cases rarely exists. More importantly, RL-based approaches decompose code generation into token-generation-action sequences, potentially restricts the reasoning capabilities of the LLMs due to the absence of a high-level thinking.

To further exploit the reasoning ability of the LLMs for better code generation, a reasonable approach is to brainstorm the blueprint of solving the programming problems in a more abstract level before actually implementing the solutions step by step. To achieve this objective, we propose the \textsc{Brainstorm} framework. The framework utilizes a \emph{brainstorming} strategy to generate diverse thoughts, i.e., high-level algorithmic blueprints of possible solutions, to help the algorithmic reasoning for code generation. Subsequently, a neural ranker model is adopted to select the best thought for solving the given problem, and finally the code generation model implements the solution based on the problem description and the selected thought. Our comprehensive experiments show that \textsc{Brainstorm} framework boosts the pass rate of ChatGPT~\cite{OpenAI2022} by a significant margin on the challenging APPS~\cite{Hendrycks2021} and CodeContests~\cite{Li2022} benchmarks, achieving the state-of-the-art performance. Furthermore, we conducted experiments on real programming contests and demonstrated that \textsc{Brainstorm} framework boost the capability to a level comparable to that of human programmers.

\begin{figure}[tbp]
\includegraphics[width=\textwidth]{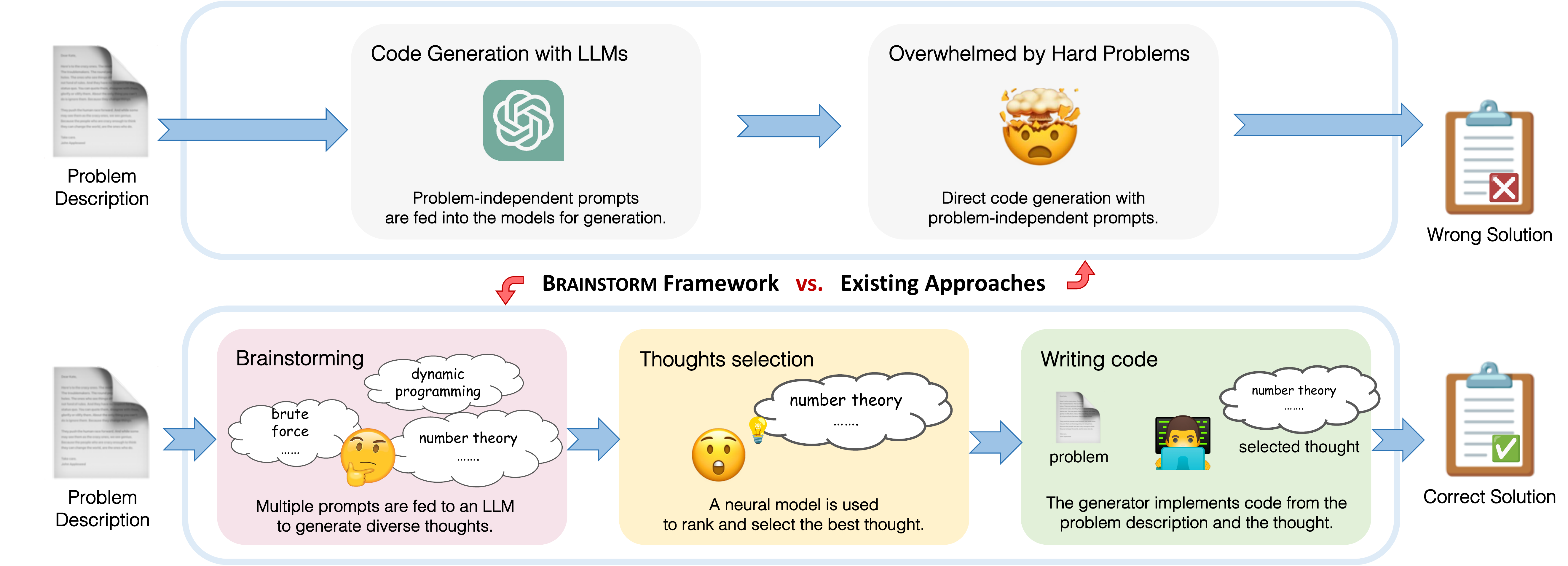}
\caption{An illustrative example for the \textsc{Brainstorm} framework.}
\label{fig:overview}
\end{figure}

The contributions of our paper lie in two folds:
\begin{enumerate}
  \item We introduce the \textsc{Brainstorm} framework, demonstrating the effectiveness of utilizing diverse ideas for competition-level code generation. By generating a variety of thoughts and selecting high-quality ones, we effectively exploit the algorithmic reasoning ability of LLMs. To the best of our knowledge, our work represents the first attempt to leverage the generation of intermediate thoughts in competition-level code generation.
  \item Our proposed approach significantly enhances the performance of LLMs in competition-level code generation. We conducted comprehensive experiments on the CodeContests and APPS benchmarks to evaluate its effectiveness. The evaluation on the CodeContests test dataset reveals substantial improvements when employing our \textsc{Brainstorm} framework. Specifically, the pass@1 metric for ChatGPT increases from 4.5\% to 7.0\%. Similarly, the pass@5 metric increases from 9.3\% to 14.7\%, while the pass@100 metric increases from 18.2\% to 29.3\%. These improvements represent relative enhancements of 55.1\%, 58.5\%, and 61.0\%, respectively. Furthermore, we observed similar improvements in the performance of the LLMs when applied to the APPS competition-level tasks.
\end{enumerate}

The remainder of this paper is organized as follows. Section~\ref{section:related_work} introduces the previous works related to our work. Section~\ref{section:framework} introduces our proposed \textsc{Brainstorm} framework. Section~\ref{section:experiments} shows that our proposed approach greatly improves the performance of LLM on the CodeContests and APPS benchmarks, and analyze the results on different settings.

\section{Related Work}\label{section:related_work}

\paragraph{Program synthesis} Program synthesis has a long history, with most classic approaches~\cite{Green1969,Manna1971,Gulwani2011} formulating the problem as searching for programs within a search space defined by the underlying programming language, where the obtained programs must satisfy the task-specific constraints. However, search-based approaches encounter the intractability of the search space and the scarcity of the formal constraints. In recent years, deep learning methods has emerged as a powerful tool for tackling these issues. Rather than searching for and validating candidates, neural approaches directly generate programs from informal specifications, such as natural language~\cite{Ling2016,Yin2017,Iyer2018,Sun2020}, partial code~\cite{Raychev2016,Murali2018,Aye2021,Guo2022}, input-output examples~\cite{Devlin2017}, or pseudo code~\cite{Kulal2019}. Despite their potential, these works are restricted to generating short programs in narrowly defined domain-specific languages, or one-line code of general-purpose programming languages.

\paragraph{Large language models} The transformer~\cite{Vaswani2017} has been successfully applied in natural language modeling~\cite{Brown2020,Devlin2019,Radford2019,Raffel2020} owing to its ability to capture long distance dependency through self-attention mechanism and its scalability. The prevailing of transformer models has created a surge of interest in using transformer in code representation learning. Prior works~\cite{Kanade2020,Feng2020,Clement2020,Guo2021,Wang2021,Lu2021,Guo2022a} have made advances in code understanding but they mostly focus on code summarization, classification and translation. Recent studies~\cite{Austin2021,Chen2021,Nijkamp2023,Xu2022,Black2022,Fried2023,Chen2023,Inala2022} investigate using pre-trained large language models for program synthesis. Codex~\cite{Chen2021} demonstrates impressive capability to complete python functions given the function signature and docstring. CodeGen~\cite{Nijkamp2023} shows that factorizing the specifications into multiple turns significantly improves synthesis quality. CodeT~\cite{Chen2023} proposes to utilize LLMs to automatically generate test cases, which effectively enhances the performance of code solution selection. Concurrently, CodeRanker~\cite{Inala2022} introduces a fault-aware ranker model that predicts the correctness of a sampled program without executing it, also aiming to address the code selection issue. However, these approaches focus on programming tasks that are no more difficult than translating instructions and invoking API calls, and as such, there is a gap with real-world programming problems.

\paragraph{Competition-level code generation} Li et al.~\cite{Li2022} proposes that current program synthesis datasets, such as HumanEval~\cite{Chen2021}, MBPP~\cite{Austin2021}, consist of mostly simple task descriptions with short solutions, which is far from the full complexity of real-world programming. To address this gap, they release CodeContests, an extensive and clone competitive programming dataset for training and evaluation. Besides, they introduce AlphaCode that is fine-tuned using GOLD~\cite{Pang2021} with tempering~\cite{Dabre2021} as the training objective and leverages large-scale sampling and filtering to obtain a small set of solutions. CodeRL~\cite{Le2022} treats the code generation model as an actor network and introduces a critic network to predict the functional correctness of generated programs, providing dense reward signals to the actor. Prompting LLMs to generate code in a zero-shot setting exhibits better potential to generalize to unseen tasks compared to fine-tuning on specific datasets. Investigations of GPT-4~\cite{OpenAI2023,Bubeck2023} demonstrate its ability to solve competition-level programming problems with a single-digit pass rate.

\begin{figure}[ht]
  \begin{minipage}{\linewidth}
   \begin{framed}\ttfamily\footnotesize
   QUESTION:\\
   {\color{gray} <problem>}\\
   Minimum LCM. Given integer n, find two positive integers a and b such that a + b = m and the least common divisor (LCM) of a and b is the minimum among all positive a and b.\\
   {\color{gray} <instructions>}\\
   Read a programming problem description on Codeforces and use your knowledge of algorithms, data structures, and mathematics to provide ideas for solving it.\\
   \\
   ANSWER:
   \end{framed}
  \centering
  (a) A prompt example for generating the diverse thoughts in brainstorming step.
  \end{minipage} \\
  
  \begin{minipage}{\linewidth}
  \begin{framed}\ttfamily\footnotesize
  We can consider a simple brute-force approach by checking all possible pairs of positive integers that add up to n and calculating their LCM.
  \end{framed}
  \centering
  (b) Example thought 1, which will lead to a \emph{time limited exceed} result.
  \end{minipage} 
  \hfill
  \begin{minipage}{\linewidth}
  \begin{framed}\ttfamily\footnotesize
  We should try to divide n as evenly as possible. If n is even, we can set a = b = n/2. If n is odd, we can set a = (n+1)/2 and b = (n-1)/2. These approaches should yield the desired result.
  \end{framed}
  \centering
  (c) Example thought 2, which will lead to a \emph{wrong answer}.
  \end{minipage} 
  \hfill
  \begin{minipage}{\linewidth}
  \begin{framed}\ttfamily\footnotesize
  We can assign the largest factor to 'a' and distribute the remaining factors to 'b'. This ensures even distribution of factors, achieving the minimum LCM among valid combinations of 'a' and 'b' that sum to 'n'.
  \end{framed}
  \centering
  (d) Example thought 3, which is a correct thought and will be selected in the thoughts selection step.
  \end{minipage} 
  
  \caption{Illustrative example for brainstorming step. Diverse thoughts are generated, the correct ones will be selected and be used for the following code generation.}
  \label{fig:brainstorm}
  \end{figure}

\section{The \textsc{Brainstorm} Framework}\label{section:framework}

Brainstorming is a frequently utilized technique among human programmers to address complex programming problems. It stimulates creative thinking and generates a diverse range of ideas aimed at discovering effective solutions. Within this variety of ideas, valuable thoughts can inspire problem reasoning. In this paper, we propose \textsc{Brainstorm} framework that integrates this human problem-solving technique into machine learning models to enhance code generation capabilities.

To be formal, let $\mathcal{P} = \{P_1,P_2,\cdots,P_{N}\}$ denote the set of problems, and $\mathcal{C} = \{C_1, C_2, \cdots , C_N\}$ denote the set of corresponding code solutions, where $N$ is the total number of problems. In the code generation task, it is natural to consider the learning of a function $f:\mathcal{P}\rightarrow\mathcal{C}$, which maps problem descriptions to their corresponding code solutions. In our \textsc{Brainstorm} framework, we propose a two-step approach to model the code generation process. The first step involves generating diverse thoughts from the problems and selecting the most suitable ones. These thoughts are denoted as $\mathcal{T}=\{T_1^1,\cdots,T_i^j,\cdots,T_N^M\}$, where $M$ represents the number of thought samples per problem. Consequently, the function $f$ is reformulated as the composition of two functions, $f = g_1 \circ g_2$. Here,~$g_1:\mathcal{P}\rightarrow\mathcal{P}\times\mathcal{T}$ maps problems to pairs of problems and thoughts, and $g_2: \mathcal{P}\times\mathcal{T}\rightarrow\mathcal{C}$ takes problem-thought pairs and produces the corresponding code solutions. And we can find that function $g_1$ and $g_2$ corresponds to the brainstorming and writing code step in the next paragraph.

As shown in Figure~\ref{fig:overview}, \textsc{Brainstorm} framework decomposes the generation task into three steps: 
\begin{enumerate}
  \item \textbf{brainstorming}: multiple prompts are constructed and fed to an LLM to generate diverse thoughts, i.e., high-level descriptions of potential solutions for the given problem.
  \item \textbf{thoughts selection}: a neural model is utilized to rank and select the thought with the highest probability of solving the given problem;
  \item \textbf{writing code}: a code generation model implements code from the problem description and the selected thought.
\end{enumerate}
The details will be introduced in the remainder of this section. It is noteworthy that our \textsc{Brainstorm} framework works in a zero-shot manner. While the performance of few-shot approaches heavily relies on the examples in the prompt, our approach excels by eliminating the need for any examples. 

\subsection{Brainstorming}
The core of \textsc{Brainstorm} framework is to generate diverse thoughts that is potentially helpful for solving the given problem. To achieve this goal, we design multiple types of instructions for the LLMs. The instructions together with the problem description are fed into the LLM to generate the diverse thoughts. The instructions used for produce thoughts are described in Appendix~\ref{appendix:instructions}.

\begin{figure}[ht]
  \includegraphics[width=\textwidth]{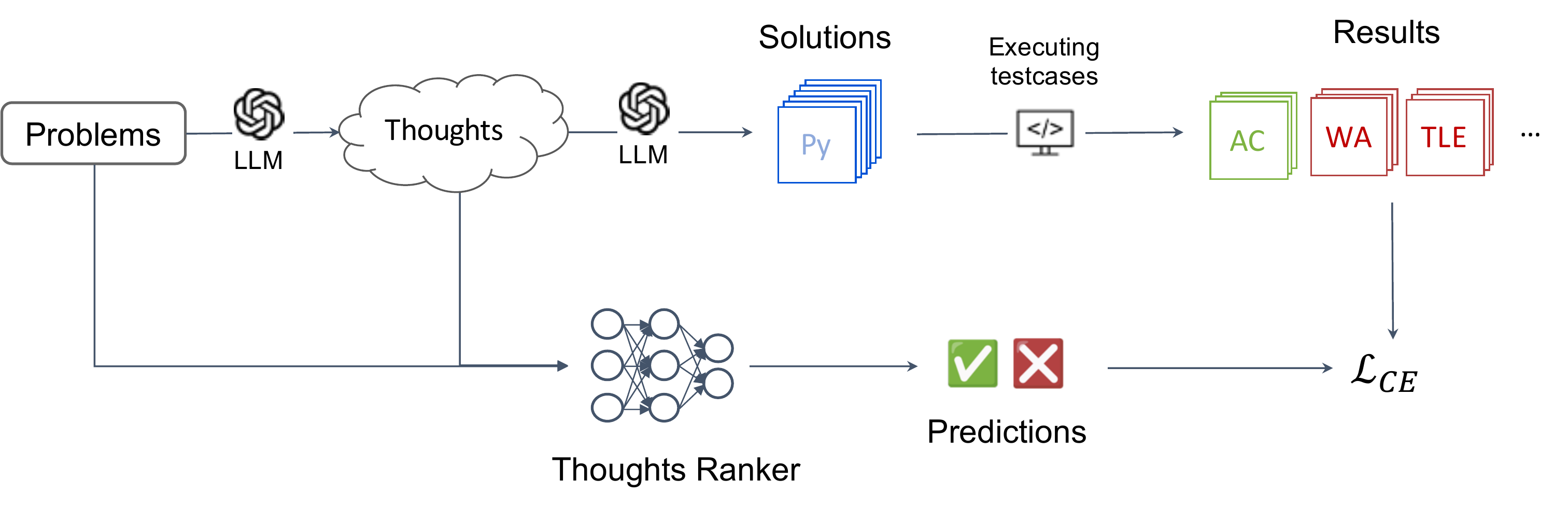}
  \caption{The overview for training thoughts ranker model.}
  \label{fig:ranker}
\end{figure}
  
As the illustrative example shown in Fig~\ref{fig:brainstorm}, brainstorming step enhances the diversity of generated algorithmic thoughts, effectively increasing the probability of correct thoughts being sampled.

Brainstorming utilizes a set of instructions to construct prompts and consequently generate diverse algorithmic thoughts, which mirrors the pattern in human reasoning: considering many possible solutions when faced with challenging problems. And we observe that the idea of brainstorming, i.e., generating intermediate thoughts and subsequently generating code, increases the diversity of sampled solutions and boosts LLMs' capability on solving complex problems.

\subsection{Thoughts Selection}

The goal of this step is to select the thoughts with a high probability of leading to the correct solution among all the candidates. To accomplish this objective, we train a neural model called the thoughts ranker, denoted as $R$. This model takes problem descriptions and the generated thoughts as input and predicts the probability that these thoughts will lead to correct solutions.

To train the thoughts ranker, we formalize the ranking as a learning task, which attempts to learn a function $r: \mathcal{P}\times\mathcal{T}\rightarrow \mathcal{Y}$. $Y_{i}^{j}\in\mathcal{Y}=\{0,1\}$ indicates whether $T_{i}^{j}$ is the correct thought that solves problem $P_i$, which can be obtained by utilizing code generation models to generate code and executing on testcases corresponding to $P_i$. The function $f$ can be learned by minimizing the following objective function

\begin{equation}
\label{eq:wec}
	\underset{r}{\min}\quad \sum_{i, j}\left[ -\lambda\ Y_i^j\log r(P_i, T_i^j) - r(P_i, T_i^j)\log Y_i^j  \right]
\end{equation}

where $\lambda$ is the hyperparameter that balances the trade-off between penalizing false positive examples and false negative examples.

Given the problem description $P_i=p_1,\dots,p_n$ and the generated thought $T_i^j=t_1,\dots,t_m$, the input sequence is constructed in the following form:  \texttt{<CLS>}, $p_1,\dots,p_n$, \texttt{<SEP>}, $t_1,\dots,t_m$, \texttt{<EOS>}. The output embedding of the \texttt{<CLS>} token is regarded as the sequence embedding and inputted to a~classification head, i.e., a~simple neural network with two linear layers and a softmax layer. 

The ranker model can be trained by minimizing the weight cross entropy described in Equation~\ref{eq:wec}. During inference, the candidate thoughts with the highest positive probability for solving problems are selected for further code generation.

\subsection{Code Generation}
After generating diverse thoughts and selecting the most suitable ones, the selected thoughts along with the problem descriptions are fed to a code generation model to implement the solution in code. The zero-shot prompt for code generation is shown in Fig~\ref{fig:zsprompt}:

\begin{figure}
\begin{lstlisting}[style=academic]
QUESTION:
<problem>						# Programming problem to solve
<thought>						# Selected thought 
Write the solution in python3.	# Instructions for code generation

ANSWER:
\end{lstlisting}
\caption{Zero-shot prompt for implementing code.}
\label{fig:zsprompt}
\end{figure}

The zero-shot prompt and the selected thought are fed to an LLM, which serves as a code generation model. The generated code is then compiled and executed on test cases in a sandbox environment to evaluate whether it can output the correct answer within the given time constraints.

\section{Evaluation}\label{section:experiments}

We next evaluate our \textsc{Brainstorm} framework with ChatGPT~\cite{OpenAI2022} as backbone large language model. We investigate (1) how \textsc{Brainstorm} framework can boost LLMs' performance on competition-level code generation benchmarks (2) the impact of two-stage generation and selector model (3) how \textsc{Brainstorm} can boost LLMs' performance on real-world programming competitions.

\subsection{Experiment Setup}

\paragraph{Benchmarks} We consider two competition-level code generation datasets for our evaluations.

1. CodeContests~\cite{Li2022}. This dataset is created for fine-tuning and evaluating AlphaCode models and consists of programming problems from a variety of sources.. The training set contains 13328 samples, the validation set 117 samples and the test set 165 samples. Li et al.~\cite{Li2022} shows that existing competitive programming benchmarks are prone to high false positive rates therefore generates extra sufficient test cases to alleviate this issue.

2. APPS~\cite{Hendrycks2021}. A dataset consists of 5000 training and 5000 test programming tasks collected from popular programming competition platforms, such as Codeforces, LeetCode\footnote{https://leetcode.com}, etc. Problems in APPS datasets are categorized into three levels according to their difficulties, introductory, interview and competition.

\paragraph{Metrics} We use pass@$k$ following the definition in Chen et al.~\cite{Chen2021}, where $k$ is set to $1,5,100$. For each programming task, $n$ samples are generated and count the number of samples $c$ which passes all test cases, the calculation of pass@$k$ is shown in the Equation~\ref{eq:passk}. Additionally, we present the pass@any metric to demonstrate the upper bound for pass@$k$ when $n$ is fixed at $200$. 

\begin{equation}
	\text{pass@}k = \underset{\text{Problems}}{\mathbb{E}}\left[1-\frac{\binom{n-c}{k}}{\binom{n}{k}}\right]
	\label{eq:passk}
\end{equation}

\paragraph{Hyperparameters} We design $m=4$ different type of instructions to prompt the LLM to generate diverse thoughts. We generate $n=200$ solutions for each programming task, which is sufficient to estimate pass@$k$ for $k\leq 100$ for both APPS and CodeContests dataset. To maintain consistency across different methods, we use the same sampling setting as Codex~\cite{Chen2021}, in which we use nucleus sampling~\cite{Holtzman2020} with top $p=0.95$ for all sampling evaluation in this work and all samples are generated with temperature $T=0.8$. 

During the training of the ranker model, we utilized the pre-trained model weights of the \texttt{xlnet-large-cased}\footnote{https://huggingface.co/xlnet-large-cased} as an initialization for our ranker model. The ranker model was fine-tuned for 8 epochs using a batch size of 32 and a learning rate of 1e-5. We used class weights to alleviate the class imbalance issue while training the ranker model. We selected the checkpoint that achieved the best pass@100 on the validation dataset. The training process was completed within 50 hours, and all experiments was conducted on 8 A100-80GB GPUs.

\paragraph{Thoughts Ranker model} One of the key obstacles in training a neural ranker model is the potential length of input sequences, as they consist of problem descriptions and generated thoughts. These input sequences often exceed the maximum input length supported by most widely used language models. To tackle this particular obstacle, we addressed the issue by implementing our ranker model through adapting the architecture of the pre-trained language model XLNet~\cite{Yang2019}. We specifically selected XLNet due to its remarkable efficacy in handling long sequences. XLNet achieves this by introducing a segment recurrence mechanism that facilitates more efficient capturing of dependencies across segments. This capability provides a clear advantage over other language models, such as BERT~\cite{Devlin2017}.

The quality of the collected corpus used for training the ranker model has a significant impact on the performance of the ranker itself and, consequently, on the overall performance of the \textsc{Brainstorm} framework. While it may seem intuitive to construct the dataset from the training split of benchmark datasets, we have observed that the presence of a domain shift between the training and test splits introduces instability and difficulty during the training process.

Appendix~\ref{appendix:dataset} provides an overview of the data sources for the CodeContests and APPS benchmarks. We assume that programming problems sourced from various origins exhibit distinct distributions, resulting in a domain shift between the training and test datasets. Re-sampling and filtering are leveraged to alleviate this issue. For the CodeContests benchmark, we specifically select programming problems from Codeforces for which \textsc{Brainstorm} generates at least one correct solution in its samples. By doing so, we construct a ranker training dataset consisting of 8k correct and 33k incorrect examples. For the APPS benchmark, we apply a filtering process to exclude data sources that do not appear in the test split. Additionally, we incorporate problems from the CodeContests benchmark to augment the insufficient data sources while taking precautions to avoid any information leakage. For example, we borrow problems from Codeforces and AtCoder\footnote{https://atcoder.jp/} in the CodeContests training set and augment the training set of the APPS benchmark with these selected problems. Similarly, programming problems for which \textsc{Brainstorm} generates at least one correct solution in its samples are selected and the ranker training dataset consists 17k correct and 64k incorrect examples.

\paragraph{Baselines} For the CodeContests benchmark, we compare our method against several baselines. These baselines include AlphaCode~\cite{Li2022} and ChatPT~\cite{OpenAI2022}.The Chain-of-Thought~\cite{Wei2022} (CoT) prompting technique significantly improves the reasoning ability of LLMs. It is worth investigating whether CoT can also enhance LLMs in the context of competition-level code generation. Therefore we include ChatGPT with CoT prompting as another baseline for comparison. Considering the potential risk of the problem description exceeding maximal input length, we employ CoT in the zero-shot setting to address this challenge.

For the APPS benchmark, our framework is compared against several baselines. These baselines can be classified into two groups based on their reliance on fine-tuning with task-specific datasets. The first group consists of fine-tuning approaches, including GPT-2~\cite{Radford2019}, GPT variants such as GPT-Neo~\cite{Black2021}, GPT-J~\cite{Wang2021a}, as well as the CodeRL~\cite{Le2022} framework with CodeT5~\cite{Wang2021} as the backbone. The second group consists of baselines within an in-context learning paradigm, including Codex~\cite{Chen2021}, Code-davinci-002, which is the latest version of Codex, and ChatGPT.

\subsection{Main Results}

\paragraph{CodeContests} For CodeContests benchmark, we evaluate the performance of \textsc{Brainstorm} on both test and validation dataset. Table \ref{table:code_contests} shows that \textsc{Brainstorm} significantly enhances the performance of ChatGPT and achieves new SOTA on both pass@$k$ and pass@any metrics. Moreover, \textsc{Brainstorm} achieves higher pass@any rate compared to AlphaCode, which involves fine-tuning a 41B pre-trained large language mode and large scale sampling, e.g.\ one hundred thousand or one million samples per problem. 

First, we analyze the results on the test dataset of 165 tasks. From Table~\ref{table:code_contests}, it can be observed that the performance of ChatGPT in conjunction with our \textsc{Brainstorm} framework exhibits a significant improvement over directly generating code on all the metrics. The pass@1 demonstrates a substantial increase from 4.52\% to 7.01\%, indicating a relative enhancement of over 50\%.

\begin{table}[ht]
  \centering
  \caption{\textbf{Results on the CodeContests test and validation dataset.} Overall, \textsc{Brainstorm} significantly boost the performance of ChatGPT on pass@$k$ metric.}
  \label{table:code_contests}
  \vspace{0.5em}
  \scriptsize
  \begin{tabular}{ccccccccc}
  \toprule
   & \multicolumn{4}{c}{Test} & \multicolumn{4}{c}{Validation} \\ \cmidrule(lr){2-5}\cmidrule(lr){6-9}
   & pass@1 & pass@5 & pass@100 & pass@any & pass@1 & pass@5 & pass@100 & pass@any \\ \midrule
  AlphaCode (fine-tuned)       & -           & -            & -            & 29.6         & -           & -            & -            & 31.4 \\
  ChatGPT                      & 4.5         & 9.3          & 18.2         & 21.2         & 6.7         & 14.8         & 24.6         & 27.4 \\
  ChatGPT + CoT                & 2.9         & 8.6          & 23.4         & 26.1         & 3.0         & 9.4          & 29.2         & 32.5 \\ \midrule
  ChatGPT + \textsc{Brainstorm}& \textbf{7.0}& \textbf{14.7}& \textbf{29.3}& \textbf{32.1}& \textbf{8.6}& \textbf{18.7}& \textbf{35.2}& \textbf{37.6} \\
  Improvement w.r.t best       & +55.1\%     & +58.5\%      & +25.29\%     & +8.5\%       & +29.1\%     & +26.7\%      & +20.6\%      & +19.8\% \\ \bottomrule
  \end{tabular}
  \end{table}
  
  \begin{table}[ht]
  \centering
  \caption{\textbf{Results on the APPS benchmark.} \textsc{Brainstorm} framework can bring performance gains of ChatGPT. The "intro", "interview", "comp" denote introductory, interview, competition-level tasks, respectively.}
  \label{table:apps}
  \vspace{0.5em}
  \scriptsize
  \setlength{\tabcolsep}{3pt}
  \begin{tabular}{ccccccccccccc}
  \toprule
  \multirow{2}{*}{Method} & \multicolumn{3}{c}{pass@1} & \multicolumn{3}{c}{pass@5} & \multicolumn{3}{c}{pass@100} & \multicolumn{3}{c}{pass@any} \\ \cmidrule(lr){2-4}\cmidrule(lr){5-7}\cmidrule(lr){8-10}\cmidrule(lr){11-13}
                               & intro         & inter         & comp         & intro         & inter         & comp          & intro         & inter         & comp          & intro         & inter         & comp \\ \midrule
  GPT-Neo                      & 3.9          & 0.6          & 0.0         & 5.5          & 0.8          & 0.0          & -            & -            & -            & 27.9         & 9.8          & 11.4 \\
  GPT-J                        & 5.6          & 1.0          & 0.5         & 9.2          & 1.7          & 1.0          & -            & -            & -            & 35.2         & 13.2         & 13.5 \\
  GPT-2                        & 1.3          & 0.7          & 0.0         & 8.8          & 3.6          & 1.0          & -            & -            & -            & 25.0         & 9.3          & 8.8 \\
  AlphaCode                    & -            & -            & -           & -            & -            & -            & -            & -            & -            & 17.7         & 5.2          & 7.1 \\
  CodeRL+CodeT5                & 7.1          & 1.9          & 0.8         & 16.4         & 5.0          & 2.8          & -            & -            & -            & 40.0         & 15.7         & 17.9 \\ \midrule
  Codex-12B                    & 4.1          & 0.1          & 0.0         & 9.7          & 0.5          & 0.1          & 20.2         & 2.0          & 1.1          & 25.0         & 3.7          & 3.2 \\
  Code-davinci-002             & 19.1         & 4.3          & 1.0         & 42.4         & 13.1         & 4.0          & 69.2         & 35.0         & 18.9         & 70.9         & 37.3         & 20.8 \\
  ChatGPT                      & \textbf{51.8}& \textbf{21.0}& 4.4         & \textbf{66.9}& \textbf{35.3}& 12.5         & 83.4         & 56.5         & 37.8         & 85.4         & 59.5         & 47.4 \\
  ChatGPT + CoT                & 42.7         & 16.0         & 3.4         & 62.6         & 30.3         & 11.5         & 81.8         & 53.9         & 39.8         & 84.5         & 57.7         & 51.7 \\ \midrule
  ChatGPT + \textsc{Brainstorm}& 46.5         & 18.1         & \textbf{5.9}& 66.1         & 33.9         & \textbf{14.1}& \textbf{85.8}& \textbf{60.3}& \textbf{52.6}& \textbf{88.3}& \textbf{64.8}& \textbf{63.8} \\
  Improvement w.r.t best       & -10.3\%      & -13.9\%      & +32.7\%     & -1.2\%       & -4.0\%       & +12.4\%      & +2.8\%       & +6.6\%       & +32.1\%      & +3.4\%       & +8.9\%       & +23.4\% \\ \bottomrule
  \end{tabular}
  \end{table}

Then, we analyze the results on the valid dataset of 117 tasks. The programming problems in the validation dataset are simpler than the ones on the test dataset, which we can see by the larger pass@$k$, and pass@any numbers. Hence, the relative improve from \textsc{Brainstorm} is smaller in scale, but it is still a significant improvement, such as the pass@1 increases by approximately 30\%.

\paragraph{APPS} For the APPS benchmark, we evaluate the performance of \textsc{Brainstorm} on a test dataset comprising 5000 tasks from various sources. Our findings indicate that \textsc{Brainstorm} improves the pass@$k$ metrics for competition-level tasks, regardless of the value of $k$, and achieves a more than 30\% relative improvement compared to the best approach when $k=1$ and $100$.

\subsection{Result Analysis}

\begin{figure}[htbp]
  \centering
  \begin{subfigure}[b]{0.48\textwidth}
    \centering
    \includegraphics[width=\textwidth]{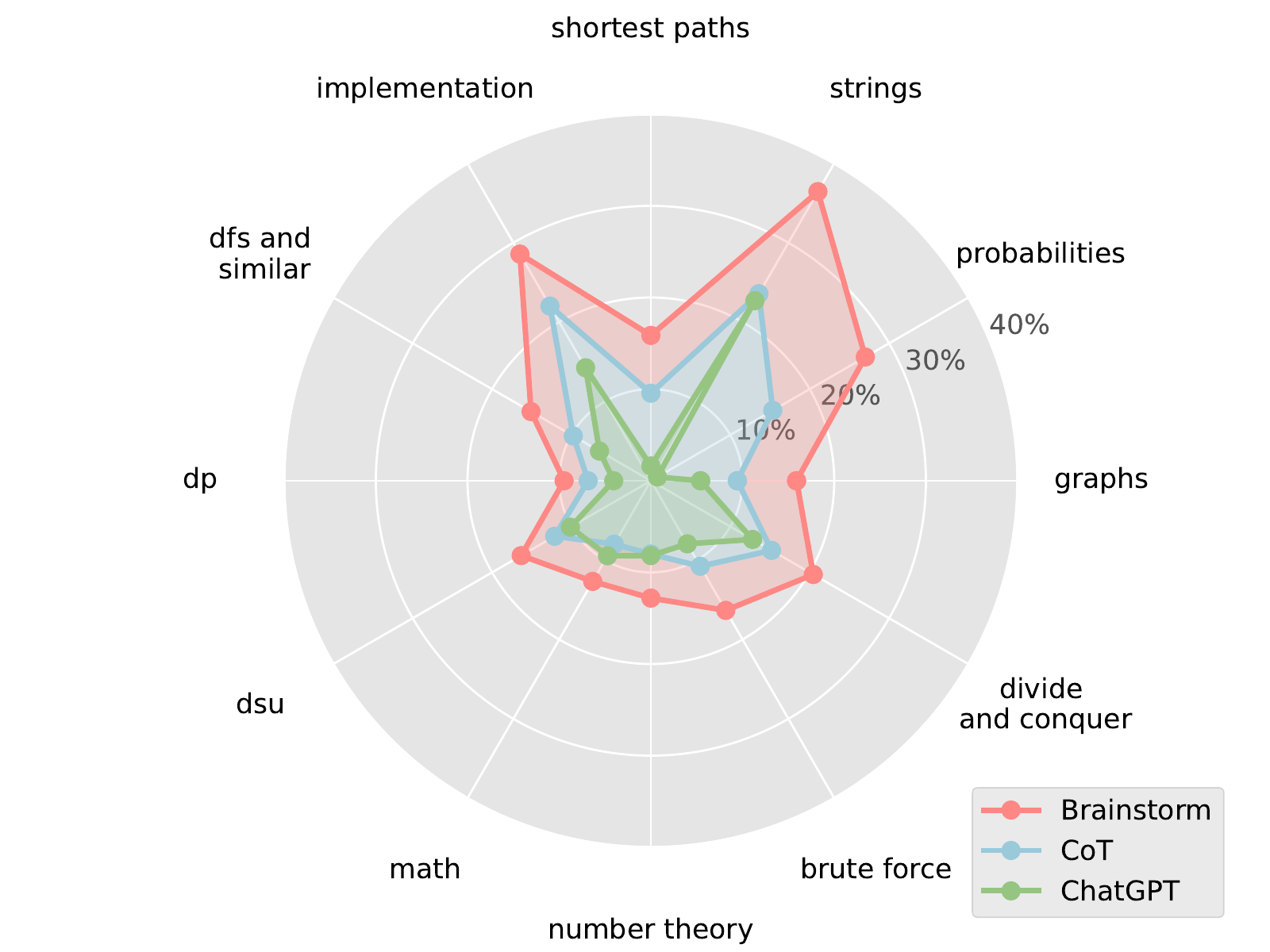}
    \caption{Pass@5 for different Codeforces Tags}
    \label{fig:subfig1}
  \end{subfigure}
  \begin{subfigure}[b]{0.48\textwidth}
    \centering
    \includegraphics[width=\textwidth]{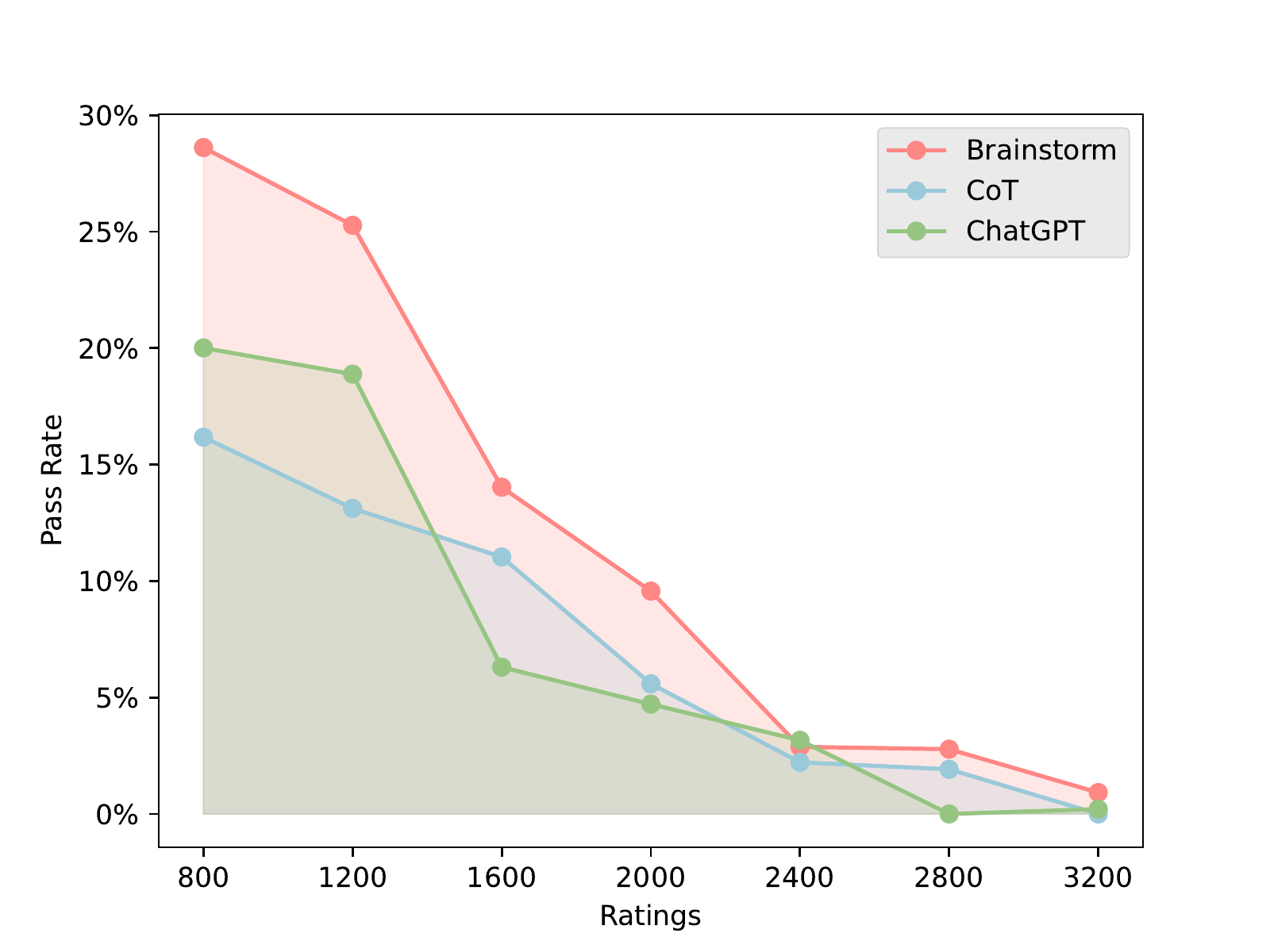}
    \caption{Pass@5 for different Codeforces Ratings}
    \label{fig:subfig2}
  \end{subfigure}
  \caption{Performance of \textsc{Brainstorm} on different tags and ratings.}
  \label{fig:combined}
\end{figure}

\paragraph{\textsc{Brainstorm} boosts algorithmic reasoning.} We conducted further analysis to examine the performance improvement of our \textsc{Brainstorm} framework on different tags and ratings within the CodeContests benchmark. As depicted in Fig~\ref{fig:subfig1}, our findings indicate a significant relative improvement in problems involving probabilities, shortest paths, and graphs. We hypothesize that this improvement can be attributed to the enhancement of algorithmic reasoning capabilities in LLMs facilitated by our \textsc{Brainstorm} framework. As illustrated in Fig~\ref{fig:subfig2}, our approach consistently outperforms ChatGPT and CoT as the rating increases. The scale of relative improvement continues to rise until the rating reaches 2000. This observation further supports the claim that \textsc{Brainstorm} boosts algorithmic reasoning, as higher-rated problems require a more comprehensive understanding of algorithms.

\paragraph{Thinking too much is harmful for simple problems.} It can be observed from Table~\ref{table:apps} that our \textsc{Brainstorm} framework exhibits improvements in all pass@$k$ metrics for competition-level tasks. However, there is a noticeable decrease in pass@1 and pass@5 for both introductory and interview-level tasks. We hypothesize that this phenomenon arises from the fact that simple problems and their corresponding solutions are well-represented in the training corpus of LLMs. As a result, ChatGPT is capable of directly solving these simple problems without generating intermediate reasoning steps. Consequently, encouraging LLMs to generate diverse thoughts through the brainstorming strategy proves to be harmful when it comes to solving simple problems.

\subsection{Performance on LeetCode Contests}

Although the \textsc{Brainstorm} framework has shown significant improvement on the competition-level code generation benchmarks, the metrics defined on these benchmarks might be prone to high false positive rates~\cite{Li2022}. To address this concern, we also evaluate our approach on the LeetCode platform, where programmers practice their skills and prepare for interviews. In order to evaluate the effectiveness of our \textsc{Brainstorm} framework, we employed the benchmark constructed by Bubeck~et~al.~\cite{Bubeck2023}, which consists of 100 problems collected from LeetCode. The generated code are submitted to the official LeetCode online judge system to check for correctness. As shown in Table~\ref{tabel:leetcode}, \textsc{Brainstorm} framework effectively boosts the ability of ChatGPT, achieving performance comparable to human.

\begin{table}[htbp]
\centering
\caption{Performance on LeetCode Contests.}
\label{tabel:leetcode}
\vspace{0.5em}
\resizebox{\textwidth}{!}{
\begin{tabular}{ccccccccc}
\toprule
\multirow{2}{*}{pass@$k$} & \multicolumn{2}{c}{Easy} & \multicolumn{2}{c}{Median} & \multicolumn{2}{c}{Hard} & \multicolumn{2}{c}{Overall} \\ \cmidrule(lr){2-3} \cmidrule(lr){4-5} \cmidrule(lr){6-7} \cmidrule(lr){8-9} 
 & $k=1$ & $k=5$ & $k=1$ & $k=5$ & $k=1$ & $k=5$ & $k=1$ & $k=5$ \\ \midrule
 Codex & 27.3 & 50.0 & 12.0 & 22.0 & 3.6 & 3.6 & 13.0 & 23.0 \\
 \texttt{text-davinci-003} & 50.0 & 81.8 & 16.0 & 34.0 & 0.0 & 3.6 & 19.0 & 36.0 \\
 ChatGPT & 50.8 & 82.5 & 13.0 & 32.1 & 0.7 & 3.8 & 19.2 & 37.3 \\
\textsc{Brainstorm} & \textbf{64.3} & \textbf{88.1} & \textbf{25.6} & \textbf{56.6} & \textbf{3.7} & \textbf{12.1} & \textbf{40.3} & \textbf{52.7} \\
\midrule
Human (Leetcode users) & \multicolumn{2}{c}{72.2} & \multicolumn{2}{c}{37.7}  & \multicolumn{2}{c}{7.0} & \multicolumn{2}{c}{38.2}  \\ \bottomrule
\end{tabular}
}
\end{table}

\section{Conclusion}\label{section:conclusion}
In this work, we present \textsc{Brainstorm}, a framework applied for competition-level code generation. Our proposed framework leverages a \emph{brainstorming} strategy to generate and select diverse thoughts that exploits the algorithmic reasoning ability of large language models, and subsequently implements the solution in a programming language. Our comprehensive experiment results demonstrate that the \textsc{Brainstorm} framework effectively promotes the diversity of generated programs and enhances ChatGPT's ability in solving competition-level programming problems.

\bibliographystyle{unsrt} \small
\bibliography{ref.bib}

\begin{thebibliography}{10}

\bibitem{Brown2020}
Tom~B. Brown, Benjamin Mann, Nick Ryder, Melanie Subbiah, Jared Kaplan,
  Prafulla Dhariwal, Arvind Neelakantan, Pranav Shyam, Girish Sastry, Amanda
  Askell, Sandhini Agarwal, Ariel Herbert{-}Voss, Gretchen Krueger, Tom
  Henighan, Rewon Child, Aditya Ramesh, Daniel~M. Ziegler, Jeffrey Wu, Clemens
  Winter, Christopher Hesse, Mark Chen, Eric Sigler, Mateusz Litwin, Scott
  Gray, Benjamin Chess, Jack Clark, Christopher Berner, Sam McCandlish, Alec
  Radford, Ilya Sutskever, and Dario Amodei.
\newblock Language models are few-shot learners.
\newblock In {\em Advances in Neural Information Processing Systems}, 2020.

\bibitem{Austin2021}
Jacob Austin, Augustus Odena, Maxwell Nye, Maarten Bosma, Henryk Michalewski,
  David Dohan, Ellen Jiang, Carrie Cai, Michael Terry, Quoc Le, et~al.
\newblock Program synthesis with large language models.
\newblock {\em arXiv preprint arXiv:2108.07732}, 2021.

\bibitem{Chen2021}
Mark Chen, Jerry Tworek, Heewoo Jun, Qiming Yuan, Henrique Ponde de~Oliveira
  Pinto, Jared Kaplan, Harri Edwards, Yuri Burda, Nicholas Joseph, Greg
  Brockman, et~al.
\newblock Evaluating large language models trained on code.
\newblock {\em arXiv preprint arXiv:2107.03374}, 2021.

\bibitem{Nijkamp2023}
Erik Nijkamp, Bo~Pang, Hiroaki Hayashi, Lifu Tu, Huan Wang, Yingbo Zhou, Silvio
  Savarese, and Caiming Xiong.
\newblock Codegen: An open large language model for code with multi-turn
  program synthesis.
\newblock In {\em Proceedings of the International Conference on Learning
  Representations}, 2023.

\bibitem{OpenAI2022}
OpenAI.
\newblock {ChatGPT}: Optimizing language models for dialogue, 2022.

\bibitem{OpenAI2023}
OpenAI.
\newblock {GPT-4} technical report.
\newblock {\em arXiv preprint arXiv:2303.08774}, 2023.

\bibitem{Li2022}
Yujia Li, David Choi, Junyoung Chung, Nate Kushman, Julian Schrittwieser,
  R{\'e}mi Leblond, Tom Eccles, James Keeling, Felix Gimeno, Agustin Dal~Lago,
  et~al.
\newblock Competition-level code generation with alphacode.
\newblock {\em Science}, 2022.

\bibitem{Zelikman2022}
Eric Zelikman, Qian Huang, Gabriel Poesia, Noah~D Goodman, and Nick Haber.
\newblock Parsel: A unified natural language framework for algorithmic
  reasoning.
\newblock {\em arXiv preprint arXiv:2212.10561}, 2022.

\bibitem{Le2022}
Hung Le, Yue Wang, Akhilesh~Deepak Gotmare, Silvio Savarese, and Steven Hoi.
\newblock {CodeRL}: Mastering code generation through pretrained models and
  deep reinforcement learning.
\newblock In {\em Advances in Neural Information Processing Systems}, 2022.

\bibitem{Hendrycks2021}
Dan Hendrycks, Steven Basart, Saurav Kadavath, Mantas Mazeika, Akul Arora,
  Ethan Guo, Collin Burns, Samir Puranik, Horace He, Dawn Song, and Jacob
  Steinhardt.
\newblock Measuring coding challenge competence with {APPS}.
\newblock In {\em Proceedings of the Neural Information Processing Systems
  Track on Datasets and Benchmarks}, 2021.

\bibitem{Green1969}
Cordell Green.
\newblock Application of theorem proving to problem solving.
\newblock In {\em Proceedings of the International Joint Conference on
  Artificial Intelligence}, 1969.

\bibitem{Manna1971}
Zohar Manna and Richard~J. Waldinger.
\newblock Toward automatic program synthesis.
\newblock {\em Commun. ACM}, 1971.

\bibitem{Gulwani2011}
Sumit Gulwani.
\newblock Automating string processing in spreadsheets using input-output
  examples.
\newblock In {\em Proceedings of the Annual ACM SIGPLAN-SIGACT Symposium on
  Principles of Programming Languages}, 2011.

\bibitem{Ling2016}
Wang Ling, Phil Blunsom, Edward Grefenstette, Karl~Moritz Hermann,
  Tom{\'a}{\v{s}} Ko{\v{c}}isk{\'y}, Fumin Wang, and Andrew Senior.
\newblock Latent predictor networks for code generation.
\newblock In {\em Proceedings of the Annual Meeting of the Association for
  Computational Linguistics}, 2016.

\bibitem{Yin2017}
Pengcheng Yin and Graham Neubig.
\newblock A syntactic neural model for general-purpose code generation.
\newblock In {\em Proceedings of the Annual Meeting of the Association for
  Computational Linguistics}, 2017.

\bibitem{Iyer2018}
Srinivasan Iyer, Ioannis Konstas, Alvin Cheung, and Luke Zettlemoyer.
\newblock Mapping language to code in programmatic context.
\newblock In {\em Proceedings of the Conference on Empirical Methods in Natural
  Language Processing}, 2018.

\bibitem{Sun2020}
Zeyu Sun, Qihao Zhu, Yingfei Xiong, Yican Sun, Lili Mou, and Lu~Zhang.
\newblock {TreeGen}: A tree-based transformer architecture for code generation.
\newblock In {\em Proceedings of the AAAI Conference on Artificial
  Intelligence}, 2020.

\bibitem{Raychev2016}
Veselin Raychev, Pavol Bielik, and Martin Vechev.
\newblock Probabilistic model for code with decision trees.
\newblock In {\em Proceedings of the ACM SIGPLAN International Conference on
  Object-Oriented Programming, Systems, Languages, and Applications}, 2016.

\bibitem{Murali2018}
Vijayaraghavan Murali, Letao Qi, Swarat Chaudhuri, and Chris Jermaine.
\newblock Neural sketch learning for conditional program generation.
\newblock In {\em Proceedings of the International Conference on Learning
  Representations}, 2018.

\bibitem{Aye2021}
Gareth~Ari Aye, Seohyun Kim, and Hongyu Li.
\newblock Learning autocompletion from real-world datasets.
\newblock In {\em Proceedings of the International Conference on Software
  Engineering: Software Engineering in Practice (ICSE-SEIP)}, 2021.

\bibitem{Guo2022}
Daya Guo, Alexey Svyatkovskiy, Jian Yin, Nan Duan, Marc Brockschmidt, and
  Miltiadis Allamanis.
\newblock Learning to complete code with sketches.
\newblock In {\em Proceedings of the International Conference on Learning
  Representations}, 2022.

\bibitem{Devlin2017}
Jacob Devlin, Jonathan Uesato, Surya Bhupatiraju, Rishabh Singh, Abdel-rahman
  Mohamed, and Pushmeet Kohli.
\newblock {RobustFill}: Neural program learning under noisy i/o.
\newblock In {\em Proceedings of the International Conference on Machine
  Learning}, 2017.

\bibitem{Kulal2019}
Sumith Kulal, Panupong Pasupat, Kartik Chandra, Mina Lee, Oded Padon, Alex
  Aiken, and Percy~S Liang.
\newblock {SPoC}: Search-based pseudocode to code.
\newblock In {\em Advances in Neural Information Processing Systems}, 2019.

\bibitem{Vaswani2017}
Ashish Vaswani, Noam Shazeer, Niki Parmar, Jakob Uszkoreit, Llion Jones,
  Aidan~N. Gomez, \L{}ukasz Kaiser, and Illia Polosukhin.
\newblock Attention is all you need.
\newblock In {\em Advances in Neural Information Processing Systems}, 2017.

\bibitem{Devlin2019}
Jacob Devlin, Ming-Wei Chang, Kenton Lee, and Kristina Toutanova.
\newblock {BERT}: Pre-training of deep bidirectional transformers for language
  understanding.
\newblock In {\em Proceedings of the Conference of the North {A}merican Chapter
  of the Association for Computational Linguistics: Human Language
  Technologies}, 2019.

\bibitem{Radford2019}
Alec Radford, Jeff Wu, Rewon Child, David Luan, Dario Amodei, and Ilya
  Sutskever.
\newblock Language models are unsupervised multitask learners.
\newblock 2019.

\bibitem{Raffel2020}
Colin Raffel, Noam Shazeer, Adam Roberts, Katherine Lee, Sharan Narang, Michael
  Matena, Yanqi Zhou, Wei Li, and Peter~J. Liu.
\newblock Exploring the limits of transfer learning with a unified text-to-text
  transformer.
\newblock {\em The Journal of Machine Learning Research}, 2020.

\bibitem{Kanade2020}
Aditya Kanade, Petros Maniatis, Gogul Balakrishnan, and Kensen Shi.
\newblock Learning and evaluating contextual embedding of source code.
\newblock In {\em Proceedings of the International Conference on Machine
  Learning}, 2020.

\bibitem{Feng2020}
Zhangyin Feng, Daya Guo, Duyu Tang, Nan Duan, Xiaocheng Feng, Ming Gong, Linjun
  Shou, Bing Qin, Ting Liu, Daxin Jiang, and Ming Zhou.
\newblock {CodeBERT}: A pre-trained model for programming and natural
  languages.
\newblock In {\em Proceedings of the Conference on Empirical Methods in Natural
  Language Processing}, 2020.

\bibitem{Clement2020}
Colin Clement, Dawn Drain, Jonathan Timcheck, Alexey Svyatkovskiy, and Neel
  Sundaresan.
\newblock {PyMT5}: Multi-mode translation of natural language and {Python} code
  with transformers.
\newblock In {\em Proceedings of the Conference on Empirical Methods in Natural
  Language Processing}, 2020.

\bibitem{Guo2021}
Daya Guo, Shuo Ren, Shuai Lu, Zhangyin Feng, Duyu Tang, Shujie LIU, Long Zhou,
  Nan Duan, Alexey Svyatkovskiy, Shengyu Fu, Michele Tufano, Shao~Kun Deng,
  Colin Clement, Dawn Drain, Neel Sundaresan, Jian Yin, Daxin Jiang, and Ming
  Zhou.
\newblock {GraphCodeBERT}: Pre-training code representations with data flow.
\newblock In {\em Proceedings of the International Conference on Learning
  Representations}, 2021.

\bibitem{Wang2021}
Yue Wang, Weishi Wang, Shafiq Joty, and Steven~C.H. Hoi.
\newblock {C}ode{T}5: Identifier-aware unified pre-trained encoder-decoder
  models for code understanding and generation.
\newblock In {\em Proceedings of the Conference on Empirical Methods in Natural
  Language Processing}, 2021.

\bibitem{Lu2021}
Shuai Lu, Daya Guo, Shuo Ren, Junjie Huang, Alexey Svyatkovskiy, Ambrosio
  Blanco, Colin Clement, Dawn Drain, Daxin Jiang, Duyu Tang, Ge~Li, Lidong
  Zhou, Linjun Shou, Long Zhou, Michele Tufano, MING GONG, Ming Zhou, Nan Duan,
  Neel Sundaresan, Shao~Kun Deng, Shengyu Fu, and Shujie LIU.
\newblock {CodeXGLUE}: A machine learning benchmark dataset for code
  understanding and generation.
\newblock In {\em Proceedings of the Neural Information Processing Systems
  Track on Datasets and Benchmarks}, 2021.

\bibitem{Guo2022a}
Daya Guo, Shuai Lu, Nan Duan, Yanlin Wang, Ming Zhou, and Jian Yin.
\newblock {UniXcoder}: Unified cross-modal pre-training for code
  representation.
\newblock In {\em Proceedings of the Annual Meeting of the Association for
  Computational Linguistics}, 2022.

\bibitem{Xu2022}
Frank~F. Xu, Uri Alon, Graham Neubig, and Vincent~Josua Hellendoorn.
\newblock A systematic evaluation of large language models of code.
\newblock In {\em Proceedings of the ACM SIGPLAN International Symposium on
  Machine Programming}, 2022.

\bibitem{Black2022}
Sid Black, Stella Biderman, Eric Hallahan, Quentin Anthony, Leo Gao, Laurence
  Golding, Horace He, Connor Leahy, Kyle McDonell, Jason Phang, et~al.
\newblock {GPT-NeoX-20B}: An open-source autoregressive language model.
\newblock {\em arXiv preprint arXiv:2204.06745}, 2022.

\bibitem{Fried2023}
Daniel Fried, Armen Aghajanyan, Jessy Lin, Sida Wang, Eric Wallace, Freda Shi,
  Ruiqi Zhong, Scott Yih, Luke Zettlemoyer, and Mike Lewis.
\newblock {InCoder}: A generative model for code infilling and synthesis.
\newblock In {\em Proceedings of the International Conference on Learning
  Representations}, 2023.

\bibitem{Chen2023}
Bei Chen, Fengji Zhang, Anh Nguyen, Daoguang Zan, Zeqi Lin, Jian-Guang Lou, and
  Weizhu Chen.
\newblock {CodeT}: Code generation with generated tests.
\newblock In {\em Proceedings of the International Conference on Learning
  Representations}, 2023.

\bibitem{Inala2022}
Jeevana~Priya Inala, Chenglong Wang, Mei Yang, Andres Codas, Mark
  Encarnaci{\'o}n, Shuvendu~K Lahiri, Madanlal Musuvathi, and Jianfeng Gao.
\newblock Fault-aware neural code rankers.
\newblock In {\em Advances in Neural Information Processing Systems}, 2022.

\bibitem{Pang2021}
Richard~Yuanzhe Pang and He~He.
\newblock Text generation by learning from demonstrations.
\newblock In {\em Proceedings of the International Conference on Learning
  Representations}, 2021.

\bibitem{Dabre2021}
Raj Dabre and Atsushi Fujita.
\newblock Investigating softmax tempering for training neural machine
  translation models.
\newblock In {\em Proceedings of Machine Translation Summit XVIII: Research
  Track}, 2021.

\bibitem{Bubeck2023}
S{\'e}bastien Bubeck, Varun Chandrasekaran, Ronen Eldan, Johannes Gehrke, Eric
  Horvitz, Ece Kamar, Peter Lee, Yin~Tat Lee, Yuanzhi Li, Scott Lundberg,
  et~al.
\newblock Sparks of artificial general intelligence: Early experiments with
  {GPT-4}.
\newblock {\em arXiv preprint arXiv:2303.12712}, 2023.

\bibitem{Holtzman2020}
Ari Holtzman, Jan Buys, Li~Du, Maxwell Forbes, and Yejin Choi.
\newblock The curious case of neural text degeneration.
\newblock In {\em Proceedings of the International Conference on Learning
  Representations}, 2020.

\bibitem{Yang2019}
Zhilin Yang, Zihang Dai, Yiming Yang, Jaime~G. Carbonell, Ruslan Salakhutdinov,
  and Quoc~V. Le.
\newblock Xlnet: Generalized autoregressive pretraining for language
  understanding.
\newblock In {\em Advances in Neural Information Processing Systems}, 2019.

\bibitem{Wei2022}
Jason Wei, Xuezhi Wang, Dale Schuurmans, Maarten Bosma, brian ichter, Fei Xia,
  Ed~H. Chi, Quoc~V Le, and Denny Zhou.
\newblock Chain of thought prompting elicits reasoning in large language
  models.
\newblock In {\em Advances in Neural Information Processing Systems}, 2022.

\bibitem{Black2021}
Sid Black, Leo Gao, Phil Wang, Connor Leahy, and Stella Biderman.
\newblock {GPT-Neo: Large Scale Autoregressive Language Modeling with
  Mesh-Tensorflow}.
\newblock 2021.

\bibitem{Wang2021a}
Ben Wang and Aran Komatsuzaki.
\newblock {GPT-J-6B: A 6 Billion Parameter Autoregressive Language Model}.
\newblock 2021.

\end{thebibliography}


\newpage
\appendix
\section{Instructions}\label{appendix:instructions}
The detailed instructions are shown in the following:

\begin{lstlisting}[style=academic, basicstyle=\ttfamily\tiny, caption={The 1st instructions.}]
Your task is to read a problem description from Codeforces and provide a detailed explanation of your approach to solving the problem without including any code. Please ensure that your explanation is clear, concise, and easy to understand for someone who may not be familiar with the specific programming language or algorithm used.

In your response, please include an overview of the problem statement and any key constraints or requirements. You should also explain how you approached the problem, including any relevant data structures, algorithms, or techniques that you used.

Please note that your response should be flexible enough to allow for various relevant and creative approaches to solving the problem.
\end{lstlisting}

\begin{lstlisting}[style=academic, basicstyle=\ttfamily\tiny, caption={The 2nd instructions.}]
Read a problem description on Codeforces and use your knowledge of algorithms, data structures, and mathematics to provide ideas for solving it. When giving an idea for solving a problem, please analyze the range of input values in detail to determine the appropriate time complexity so as to avoid timeout errors. Please note that your answers should not contain any form of code or programming language.

To make your problem-solving ideas more creative and unique, be sure to fully explain the algorithms, data structures, and mathematical concepts involved. At the same time, when discussing time complexity, please explain in as much detail as possible how to derive a feasible time complexity based on the range of input values.

\end{lstlisting}

\begin{lstlisting}[style=academic, basicstyle=\ttfamily\tiny, caption={The 3rd instructions.}]
Your task is to analyze a problem description from Codeforces and provide a solution strategy that utilizes appropriate algorithms, data structures, and mathematical concepts. You should consider the input range of the problem to determine an achievable time complexity for your solution and avoid runtime errors.

Please provide a clear explanation of your thought process in developing the solution strategy, including any relevant formulas or equations used. Your response should focus on the underlying principles behind the algorithm rather than providing specific code implementations.

Note that you are not required to include any actual code in your response.

The problem is described as follows:

\end{lstlisting}

\begin{lstlisting}[style=academic, basicstyle=\ttfamily\tiny, stringstyle=\color{black}, caption={The 4th instructions.}]
Suppose you are a programming teacher and you will given high-level thoughts after reading the problem description from codeforces. 
1. Thoughts should be written in natural language and not include any form of code or pseudo-code.
2. Thoughts should not include any reference or to external resources.
3. We prefer the simple solution if the problem has multiple solutions.
4. Priorities from high to low: brute-force, greedy, dynamic programming ...

Let's think step by step and come up with a clever and efficient solution. 
\end{lstlisting}

\section{Ranker datasets}\label{appendix:dataset}

\begin{table}[htbp]
    \centering
    \caption{Data sources of the APPS and CodeContests benchmarks.}
    \label{table:sources}
    \resizebox{0.8\textwidth}{!}{
		\begin{tabular}{cccccccc}
		\toprule
 			&  & Codeforces & Aizu & AtCoder & HackerEarch & CodeChef & \\ \midrule
			\multirow{3}{*}{CodeContests} & training & 7819 & 2161 & 1323 & 1247 & 768 &\\
 			& validation & 117 & - & - & - & - \\
 			& test & 165 & - & - & - & - \\ \bottomrule
 			\noalign{\vskip 2mm}
			\toprule
 			&  & Codeforces & CodeWars & AtCoder & Open Kattis  & CodeChef &\\ \midrule
			\multirow{2}{*}{APPS} & training & 477 & 2515 & 61 & - & 1112 &\\
 			& test & 2953 & - & 696 & 1226 & 61 & \\ \bottomrule
		\end{tabular}
	}
\end{table}

\end{document}